\documentclass[letterpaper, 10 pt, conference]{ieeeconf}
\IEEEoverridecommandlockouts
\usepackage{import}
\usepackage{url}
\usepackage{verbatim}
\usepackage{latexsym}
\usepackage{cite}
\usepackage{caption}
\usepackage{xspace}
\usepackage{hyperref}
\usepackage[letterpaper, portrait, margin=1in]{geometry}
\usepackage{float}
\usepackage{dirtytalk}
\usepackage{fullpage}
\usepackage{times}
\usepackage{fancyhdr,graphicx,amsmath,amssymb}
\usepackage[ruled,vlined]{algorithm2e}
\usepackage{tabu}
\usepackage{soul}
\usepackage{tabularx}
\usepackage{balance}
\usepackage{multicol}
\usepackage[table,xcdraw]{xcolor}

\title{\LARGE \bf Adaptive Lookahead Pure-Pursuit for Autonomous Racing}

\author{Varundev Sukhil \& Madhur Behl \\
Dept. of Computer Science, University of Virginia \\
\{varundev, madhur.behl\}@virginia.edu}

\begin{document}

\maketitle

\begin{abstract}
This paper presents an adaptive lookahead pure-pursuit lateral controller for optimizing racing metrics such as lap time, average lap speed, and deviation from a reference trajectory in an autonomous racing scenario. 
We propose a greedy algorithm to compute and assign optimal lookahead distances for the pure-pursuit controller for each waypoint on a reference trajectory for improving the race metrics.
We use a ROS based autonomous racing simulator to evaluate the adaptive pure-pursuit algorithm and compare our method with several other pure-pursuit based lateral controllers.
We also demonstrate our approach on a scaled real testbed using a F1/10 autonomous racecar.
Our method results in a significant improvement ($20\%$) in the racing metrics for an autonomous racecar. 
\end{abstract}

\section{Introduction}


Autonomous racing can be considered as the extreme version (high speeds, and close proximity to other self-driving agents) of the self-driving car problem, and therefore making progress here will enable breakthroughs in agile, and safe autonomy.
Autonomous racing is already becoming a futuristic motor-sport~\cite{scacchi2018autonomous}. 
Roborace~\cite{roborace} is the Formula E's sister series, which features fully autonomous race cars.
International autonomous racing competitions such as F1/10 autonomous racing~\cite{f1tenth, o2019f1}, Autonomous Formula SAE~\cite{koppula2017learning} are becoming proving grounds for testing the perception, planing, and control algorithms at higher speeds. 
Amazon has also recently announced a 1/18 scale DeepRacer testbed~\cite{amazon} for end-to-end driving and reinforcement learning methods for autonomous racing.

For a single vehicle to race autonomously around a track, the environment around the car on the racetrack must be perceived. This is typically done using a Simultaneous Localization And Mapping (SLAM) algorithm (\cite{durrant2006simultaneous,montemerlo2002fastslam,bailey2006simultaneous,hess2016real}). Next, the map is used to obtain a reference trajectory (\cite{krogh1986integrated,shiller1991dynamic,frazzoli2002real}) that the race car can follow. 
Finally, the vehicle's steering and velocity controller is fed with small trajectory parts with a defined time horizon while the car is driving around the track. 

This combination of path planning and motion control is a critical capability for autonomous vehicles. Pure-pursuit controllers are a prevalent class of geometric lateral control algorithms for steering autonomous cars. 
This paper focuses on advancing the design of an adaptive version of the Ackermann-adjusted pure-pursuit controller~\cite{park2015development} to make it suitable for the purpose of autonomous racing. 
The analysis and scope of this paper is limited to the single agent setting, where a single autonomous race car is tasked with following a reference trajectory (often the raceline), with the minimum lap time.
This is known as the \emph{time-trial} racing problem. 

\noindent \textbf{Research contributions of this paper}:
With the autonomous racing time-trial scenario in mind, this paper has the following novel contributions:

\begin{enumerate}
\item A greedy algorithm for adaptive lookahead pure-pursuit: given a reference trajectory, our offline algorithm produces the optimal lookahead distance assignment for each waypoint on the reference trajectory based on a tunable convex racing objective.
\item We demonstrate the increased performance in lap time and average speed of the adaptive lookahead pure-pursuit implementations and compare them to a baseline Ackermann-adjusted pure-pursuit in a Gazebo based racing simulator\cite{f1tenthsim} \& on a real scaled F1/10 autonomous racecar\cite{o2019f1}.
\end{enumerate}

\section{Related Work}

Autonomous racing has received attention in recent years from the robotics, control systems, autonomous vehicles, and deep learning communities. 
In \cite{liniger2015optimization}, authors present the use of nonlinear model predictive controller (NMPC) for the control of 1:43 scale RC race cars. Using a dynamical model of the vehicle, the authors compute racing trajectories and control inputs using receding horizon based controller.
A similar MPC based controller is also presented in~\cite{rosolia2017autonomous,hoffmann2007autonomous}.
In~\cite{ni2017dynamics}, authors design a controller to drive an autonomous vehicle at the limits of its control and traction.
AutoRally, an open-source 1:5 scale vehicle platform for aggressive autonomous diving is presented in~\cite{goldfain2018autorally}.
In all of this work, the MPC directly generates the steering and throttle control inputs based on the reference trajectory and the state of the vehicle. With these approaches an accurate and detailed dynamical model is required. 

Researchers have also analyzed the problem of computing the optimal (fastest) raceline for a given track layout.
A minimum curvature trajectory controller for the Roborace DevBot autonomous racecar is described in~\cite{doi:10.1080/00423114.2019.1631455}.
\cite{kelly2010time, tipping2011racing, xiong2010racing,cardamone2009line, theodosis2013nonlinear} addresses the problem of computing the optimal racing line. 
In our work, we assume that the race line is known a-priori and provided as a reference trajectory.
Our proposed adaptive lookahead pure-pursuit algorithm can work for any reference trajectory. 

Path tracking is the problem concerned with determining speed and steering inputs at each instant of time in order for the robot to follow a certain path. 
In~\cite{doi:10.1177/1077546316646906}, authors describe a model-based receding horizon controller for pure-pursuit tracking. They accommodate the vehicle’s steady-state lateral dynamics to improve tracking performance at high speeds. 
\cite{morales2009pure} investigates the application of the pure-pursuit technique for reactive tracking of paths for nonholonomic mobile robots. 
Researchers have also analyzed the stability of mobile robot path tracing algorithms~\cite{ollero1995stability} including pure-pursuit. 
We guide the reader towards~\cite{samuel2016review} for a detailed review of the applications of pure-pursuit.

Previous work on overcoming the limitations of pure-pursuit like corner cutting and limited maximum speed are addressed in~\cite{corner_cutting_1, corner_cutting_2} and have been successful within the stated scope of those projects. However, the metrics for an autonomous racecar as defined in this paper require a different approach which addresses a combination of the previous work and a novel method to maximize a global racing objective.

\section{Problem Formulation}
We present a brief overview of the pure-pursuit algorithm in order to provide the background and motivation for our work on adaptive lookahead pure-pursuit. 

\subsection{Pure-Pursuit Algorithm}
\label{subsec:ppa}

Pure-pursuit is a seminal algorithm for geometric lateral control that can be easily implemented in several applications including autonomous robots. 
It can  be  dated  back  in  history  to  the  pursuit  of missile  to  a  target~\cite{scharf1969comparison}.  
This algorithm is popular for it's ability to recover if the robot moves too far away from the reference trajectory. 

\noindent \textbf{Seminal Pure-Pursuit}

Pure-pursuit computes the angular velocity command that moves the robot from its current position to reach a lookahead waypoint in front of the robot. 
The linear velocity is assumed constant.
As the robot pursues the goal, the algorithm then moves the lookahead point further on the path based on the current position of the robot.
The original pure-pursuit algorithm~\cite{coulter1992implementation} was implemented on full-differential drive robot while taking into account it's associated kino-dynamic constraints. 

Consider a robot $R$ whose pose is $(x_1,y_1,\phi)$ where $(x_1,y_1)$ represent the 2D position of the robot and $\phi$ is it's current heading in the local frame, and a goal position ($x_2,y_2$) that is lookahead distance $l_d$ away on the reference trajectory. 
The pure-pursuit controller is tasked with finding the curvature of the circular arc that will guide the robot from it's current position to the goal. The relative angular offset, $\alpha$, between the robot's current heading and the goal, and the curvature $k$ is calculated using:

\begin{equation}
\alpha = \tan^{-1}(\frac{y_2-y_1}{x_2-x_1}); \quad k = \frac{2 \sin(\alpha)}{l_d}
\end{equation}



\noindent The curvature provided by equation (2) is used to calculate the  heading required to move the robot at a constant speed along the circular arc.
Once the arc is computed, the robot follows the arc at a fixed velocity for a certain time $\tau$, before recomputing the goal based on the lookahead distance. 


The \texttt{LookAheadDistance}, $l_d$ parameter controls how far along the reference path the robot should look from the current location to compute the steering/lateral correction commands. 
Changing this parameter affects the tracking behaviour of the robot: if the distance is low, it can lead to oscillations around the reference path, and if it is too high, it can cause large deviations and lead to corner-cutting~\cite{corner_cutting_1, corner_cutting_2}.

\begin{figure}[H]
    \centering
    \includegraphics[width=\linewidth]{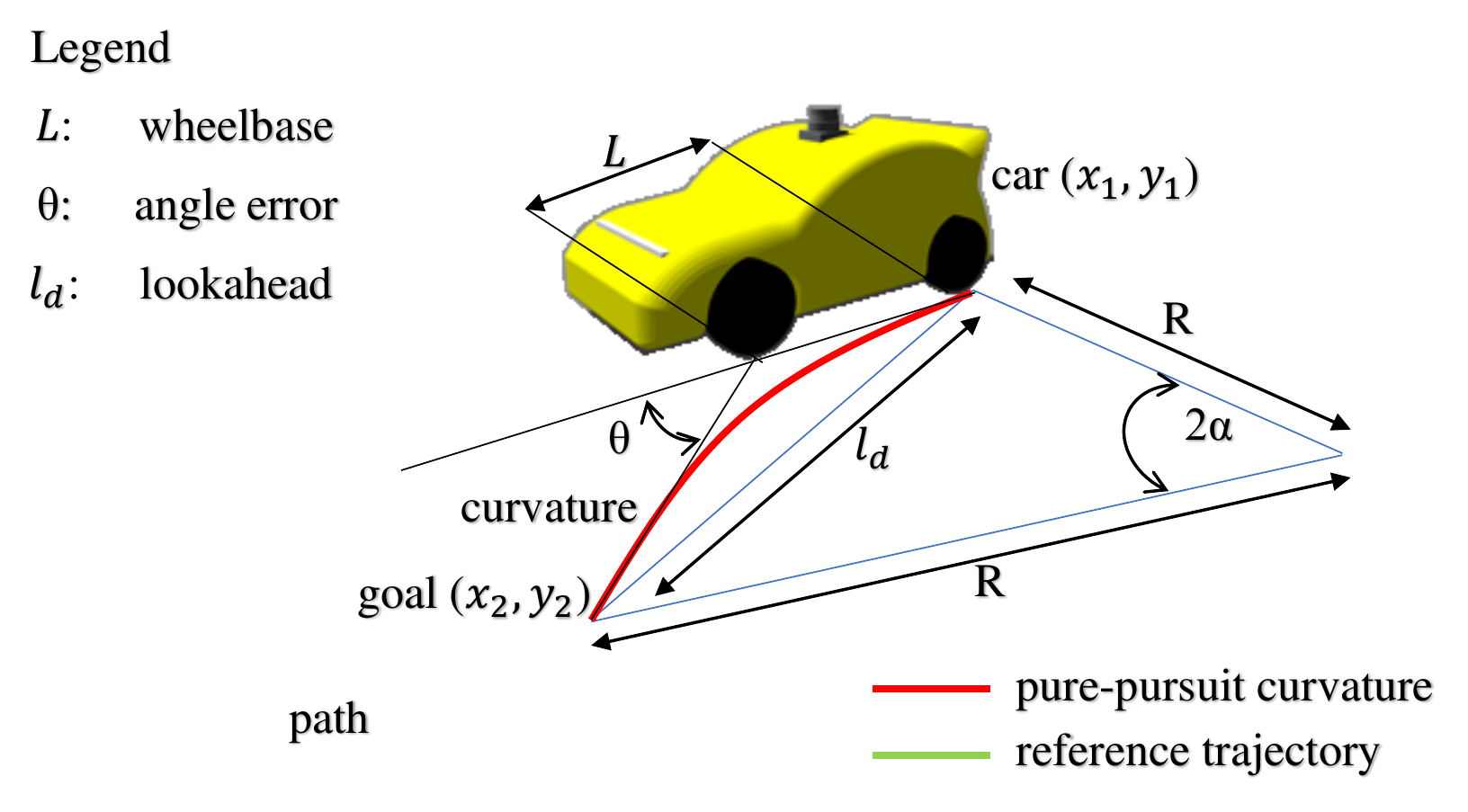}
    \caption{Calculating the desired heading $\theta$ using Ackermann-adjusted pure-pursuit from the racecar's $base\_link$ at the center of the rear axle}
    \label{fig:fig_1}
\end{figure}

\noindent \textbf{Ackermann-Steering Adjustment}

The seminal pure-pursuit produces undesired driving behavior like cutting corners~\cite{chen1999experimental} when implemented in an Ackermann-steering~\cite{mitchell2006analysis} robot and when the look-ahead parameter is not well tuned~\cite{campbell2007steering}.
For implementing pure-pursuit path tracking controller to non-holonomic Ackermann-steering robots, we need to add the geometric constraints of the robot to equation (1). 
To do so, we use the Ackermann adjusted pure-pursuit implementation as described in~\cite{park2015development}.
We define the $base\_link$ ($x_1,y_1$) as the center of the rear-axle of the racecar. 
By including the robot wheelbase $L$ (distance between the front and the rear axle), the pure-pursuit controller calculates the heading $\theta$ required to guide the robot along the curvature as:

\begin{equation}
\theta = tan^{-1}(kL) = tan^{-1}(\frac{2L sin(\alpha)}{l_d})
\end{equation}

This is depicted in Figure~\ref{fig:fig_1}. 
The racecar finds the nearest point to its $base\_link$ in the reference trajectory and identifies a goal waypoint on the trajectory that is distance $l_d$ away from the $base\_link$. It then computes the arc of radius $R$ that joins the $base\_link$ to the goal to find the angular offset $\alpha$. 
Adjusting for the racecar's wheelbase $L$, the angular offset to the goal is calculated with reference to the front axle that is distance $L$ away from the $base\_link$ in the heading of the racecar. This heading is $\theta$, and it is calculated from equation (2).
The curvature $k$, goal, ($x_2,y_2$) and angle $\theta$ are continuously updated as the racecar follows the reference trajectory.

\subsection{Autonomous racing problem setup}
We define a race-track as any closed-loop drivable environment.
The reference trajectory is a sequence of way-points that the car can follow.
As described earlier, there are several ways of choosing the right reference trajectory - mathematical race lines such as minimum distance, or minimum curvature - or complicated race lines computed while taking into account the dynamics of the vehicle.

Let $\mathcal{W}$ denote the set of $N$ waypoints $w_i$ that collectively form the reference trajectory:
\begin{equation}
\mathcal{W} = \{w_i, \quad i \quad \epsilon \quad 1 \rightarrow N\}
\end{equation}

Each waypoint $w_i$, represents the coordinates $(x\_map_i,y\_map_i)$ from the beginning of the start, and heading $\theta_i$ to the next waypoint $w_{i+1}$, i.e.

\begin{equation}
w_i = \{x\_map_i, y\_map_i, \theta_i\}
\end{equation}

Our approach is agnostic to whether the reference trajectory is optimal or not, and will work as long as any reference trajectory is a closed-loop. 

\subsection{Adaptive Pure-Pursuit Problem Statement}

In racing, the ultimate objective is to be faster than your opponents. This can be translated into having a lower lap time than the opponents. 
The lap time depends on many factors, including average velocity around the track, total distance travelled etc. 
In the absence of other opponents, the goal is to stick to the reference trajectory and be as fast as possible. 
For this paper we assume a single racecar on the track at any time (time-trial mode).

As described in Section~\ref{subsec:ppa}, the lookahead distance $l_d$ of the pure-pursuit controller is the most important parameter which determines the behavior of the autonomous racecar.
We pose the following question: 

\noindent \emph{What is the optimal value of the lookahead distance for a pure-pursuit controller that will result in the fastest lap around the track ?}

One can think of this as an offline label assignment problem, where we want to assign each waypoint $w_i$, on the reference trajectory an associated optimal lookahead distance, $l_j$ that the pure-pursuit controller will take as input when it arrives at that waypoint. 
This idea forms the basis for an \emph{adaptive lookahead pure-pursuit} controller.  

Given a reference trajectory $\mathcal{W}$ that consists of way-points described in equation (4), consider a set of $K$ lookahead distances (labels) $\mathcal{L}$: 

\begin{equation}
\mathcal{L} = \{l_j, \quad j \quad \epsilon \quad 1 \dots K\}
\end{equation}

A lookahead label informs the underlying pure-pursuit controller about the control horizon. The pose of the racecar at any given time in the race-track is denoted as the tuple $\mathcal{T}_i$, such that;

\begin{equation}
\mathcal{T}_{i} = <x_i,y_i,\phi_i,v_i>
\end{equation}

Where $(x_i,y_i)$ is the position of the racecar in the race-track relative to the start/finish line, $\phi_i$ is the heading of the racecar at the given position, and $v_i$ is the current velocity of the racecar. 
Given each way-point $w_i$ and the lookahead distance set $\mathcal{L}$ we want to compute a function assignment $\gamma$ such that;

\begin{equation}
\gamma(w_i,l_j,T_i) \rightarrow (v_{exit\_i}, \delta_i)
\label{eq:gamma}
\end{equation}

Where $v_{exit\_i}$ is the exit velocity of the racecar, and $\delta_i$ is the deviation from the reference trajectory of the racecar for the given way-point and lookahead distance. 

The label assignment \textit{policy} $\pi$ can be defined as a mapping for every way-point with an optimal lookahead distance from the set $\mathcal{L}$.

\section{Adaptive Lookahead Pure-Pursuit}

\begin{algorithm}
\SetAlgoLined
\textbf{Input}: $T_{init}, v_{init} = 0, \mathcal{W}, \mathcal{L}$ \\
\textbf{Compute}: \\
\While{$i < N$}{
    $R = SpawnCar(T_i)$ \\
    \For{$l_j = 0$ \KwTo $K$}
        {
        PurePursuit($l_j$, $R$) \\
        \eIf{CrashDetected()}{
        ResetCar($R$) \\
        $v_{exit\_i} = 0$ \\
        $\delta_i = \infty $ \\ }
        {
        \textbf{until} $R = l_j|\mathcal{W}$ \\
        calculate $\{v_{current}, \delta_{current}\}$ \\
        \textbf{then}: \\
        $v_{exit\_i} = v_{current}$ \\
        $\delta_i = \delta_{current}$ \\
        $v_{i+1} = v_{exit\_i}$ }
    }
    $\pi(w_i) = \pi^*(\beta, v_{exit\_i}, \delta_i) \forall l_j \epsilon L$
}
\KwResult{$\pi = l_i \forall w_i; l_i \epsilon L$}
\caption{Lookahead label Assignment}
\label{alg:gpp}
\end{algorithm}

Race-tracks have sections of lengthy corridors with no turns or small angled turns, and a racecar must utilize these sections of the race-track to achieve higher speeds in order to minimize lap times. 
Longer lookahead distances yield higher speeds; but at tight turns, the racecar will attempt to cut corners leading to collisions with the bounds of the race-track. 
This means the the lookahead distance has to be tuned to work with the most difficult section of the race-track.
Consequentially, we use multiple lookahead distances for different sections of the track. 
We find the labelling \textit{policy} $\pi$ which assigns lookahead distances to different sections of the track based on the desired racing objectives.


We first define the racing objectives:
\subsubsection{Maximum Velocity Pure-Pursuit (vel*)}

The labelling \textit{policy} $\pi_v^*$ maximizes the $v_{exit\_i}$ exit velocity for each way-point $w_i$ by selecting appropriate the lookahead distance $l_i$ from the set $\mathcal{L}$, i.e.
\begin{equation}
\pi_v^* = arg \max_{\mathcal{L}}(\sum_{i=1}^{N} v_{exit\_i}) \quad \forall \quad l_i \quad \epsilon \quad (1 \dots K)
\label{eq:vel_greedy}
\end{equation}

\subsubsection{Minimum Deviation Pure-Pursuit (dev*)}

The labelling \textit{policy} $\pi_\delta^*$ minimizes $\delta_i$ deviation for each waypoint $w_i$ by selecting the lookahead distance $l_i$ which produces the minimum $\delta_i$ for all looaheads  in $\mathcal{L}$. We define deviation as the area of the curve between the reference trajectory and the actual trajectory taken by the racecar.

\begin{equation}
\pi_\delta^* = arg \min_{\mathcal{L}}(\sum_{i=1}^{N} \delta_i) \quad \forall \quad l_i \quad \epsilon \quad (1 \dots K)
\label{eq:dis_greedy}
\end{equation}

\subsubsection{Convex Combination Pure-Pursuit}

This objective is a convex combination of previous two objectives from equations (8) and (9), governed by trade-off factor $\beta$. 
\begin{equation}
\pi_{v-\delta}^* = \beta(\pi_v^*) + (1 - \beta)(\pi_\delta^*); \quad \beta \quad \epsilon \quad [0,1]
\label{eq:con_greedy}
\end{equation}

\begin{figure}[H]
    \centering
    \includegraphics[width=0.8\linewidth]{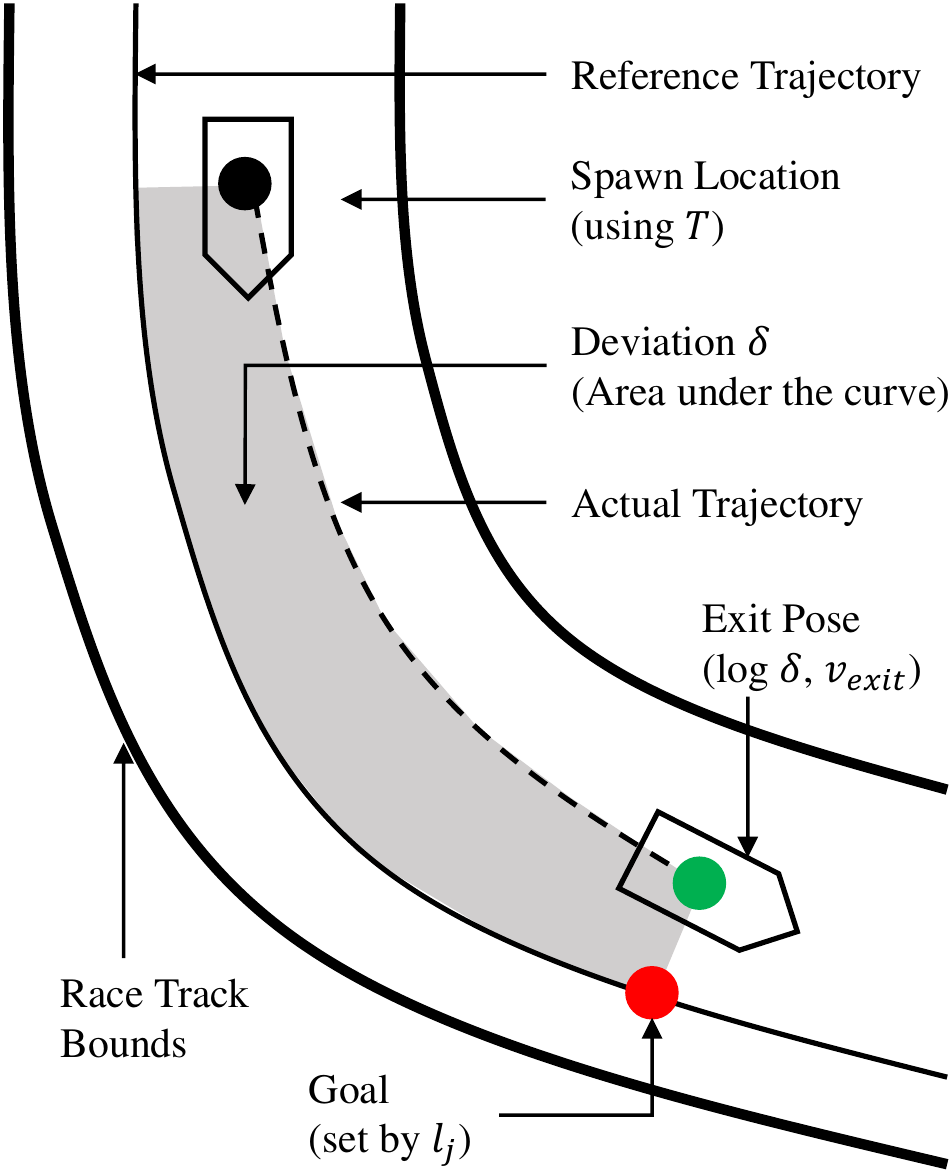}
    \caption{An iteration of the lookahead label assignment algorithm, with racecar spawned using $\mathcal{T}$, the goal set by current lookahead $l_j$, and the actual trajectory taken by the racecar using the current lookahead until the goal - where the exit pose, deviation ($\delta$) and $v_{exit}$ are logged}
    \label{fig:fig_algo}
\end{figure}

Depending on the application, trade-off factor $\beta$ can be adjusted such that $\beta = 0$ produces minimum deviation and $\beta = 1$ produces maximum achievable velocity.

Having defined the different objectives for the adaptive lookahead label assignment, we now present a novel lookahead label algorithm which can assign the optimal lookahead distance labels to each way-point based on the specified objective function (Eqs~\ref{eq:vel_greedy},\ref{eq:dis_greedy},\& \ref{eq:con_greedy}).
An overview of our method is presented in Algo.~\ref{alg:gpp}, and a visual representation is provided in Fig.~\ref{fig:fig_algo}.

\begin{figure*}[h!]
    \centering
    \includegraphics[width=0.92\linewidth]{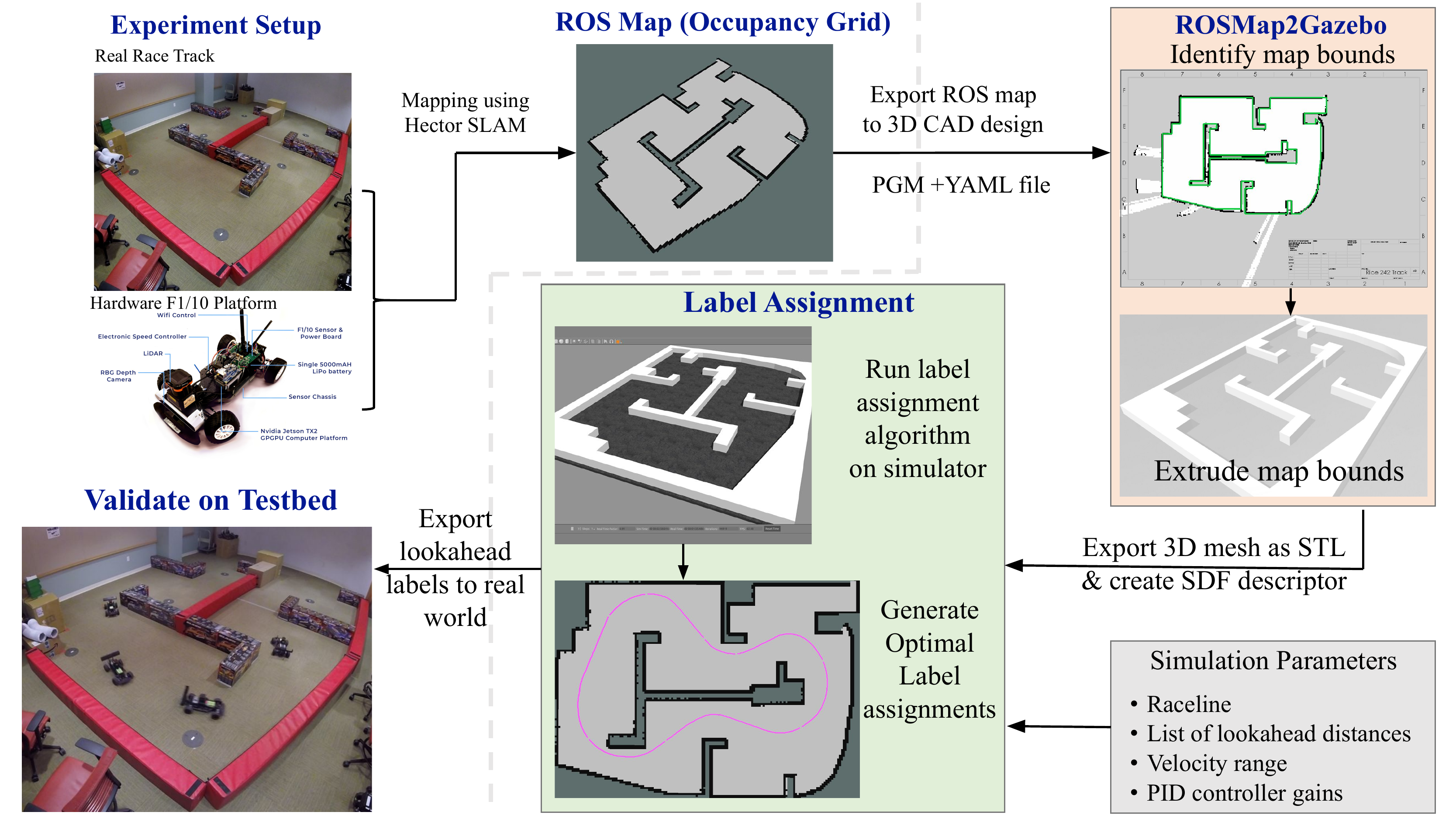}
    \caption{Clockwise from Top Left: The F1/10 platform is manually driven around the race track to create a ROS map using traditional SLAM, the ROS map is exported to CAD where the map bounds are extruded and exported as a 3D mesh, the label assignment algorithm is performed on the new map using the set simulation parameters, \& the labels are exported to be validated on the F1/10 platform}
    \label{fig:realworld}
\end{figure*}

For a waypoint $w_i$, we spawn the autonomous car in the simulator using the tuple $\mathcal{T}_i$, using the function $SpawnCar()$. 
At this waypoint, we simulate the function $\gamma$ (Eq~\ref{eq:gamma}) for each of the possible lookahead value in the set $\mathcal{L}$. 
Each iteration of $\gamma$ makes the racecar use Ackermann-adjusted pure-pursuit using the current lookahead until it approaches the goal on the reference trajectory originally set when the racecar was spawned (at both spawn and goal, the Euclidean distance between the racecar's $base\_link$ and the actual corresponding waypoint on the reference trajectory is minimal compared to all other waypoints in $\mathcal{W}$), during which time the algorithm continuously computes the racecar's deviation from the reference trajectory and its current velocity. At the end of the current iteration, when the racecar is closest to the original goal set at spawn, the exit velocity, $v_{exit\_i}$, which is the current velocity when the racecar is closest to the original goal, and the total deviation from the reference trajectory, $\delta_i$, from when the racecar travelled from the spawn location to the goal is logged.
If the racecar collides with the race-track boundaries at any time during the current iteration of the algorithm, the corresponding lookahead distances are not considered as candidates for selection at the current waypoint $w_i$. This is captured by the $CrashDetected()$ subroutine in Algorithm~\ref{alg:gpp}.
When the algorithm completely iterates through all lookaheads in $\mathcal{L}$ for all waypoints in $\mathcal{W}$, the logged data which contains [$v_{exit_i}$, $\delta_i$] is match to the corresponding lookahead and stored for offline tuning.

Next, we greedily select the lookahead distance which is best suited for the objective function using the policy $\pi^*$. For e.g. for $vel*$, we would pick the lookahead distance with the maximum exit velocity at each waypoint and the corresponding lookahead is assigned as its label. 
The same criteria can be applied to the $dev*$, and $convex\_combination$ objectives. The policy is applied to assign the best corresponding lookahead label from $\mathcal{L}$ for all waypoints in $\mathcal{W}$.


\begin{figure*}[h!]
    \centering
    \includegraphics[width=0.92\linewidth]{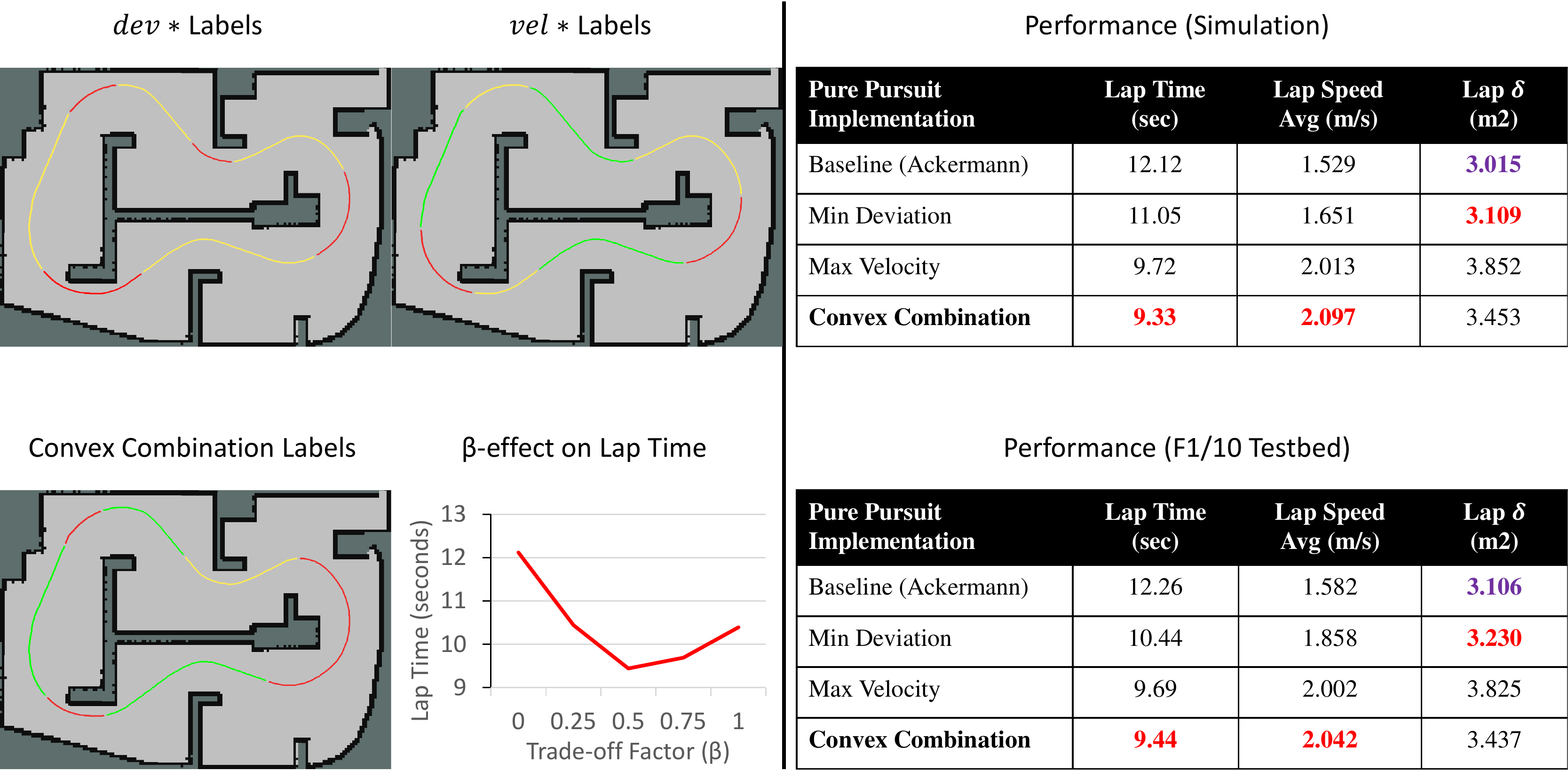}
    \caption{[Left Half]: The labels generated using the lookahead label assignment assignment for various values of $\beta$; [Right Half]: Race metrics performance of the various pure-pursuit implementations compared to the baseline Ackermann-adjusted pure-pursuit on the simulator and testbed}
    \label{fig:results_graphs}
\end{figure*}



\label{sec:alpp}

\section{Implementation on Simulator \& Testbed}


The race-track used in the experiment is a small indoor setup with tight turns, and to ease computation on the racecar's onboard embedded computer (we use the NVIDIA Jetson TX2), we decided to limit the number of lookaheads to 3. While this is not a limitation of the algorithm, the observable differences in performance of the racecar at tightly grouped lookaheads did not produce a larger racing performance increase compared to the additional computation demanded by the onboard computer. We chose lookahead distances (set $ \mathcal{L} = \{1.0, 1.5, 2.0\}, \quad  K = 3$). Empirically, the racecar tracked the reference trajectory best at  $1.0m$ lookahead, and at $2.0m$ lookahead, the racecar was able to achieve the maximum permissible velocity.


\subsection{Experiment Setup}

Fig.~\ref{fig:realworld} provides an overview of the experiment workflow, where the major steps are descirbed below:
\begin{enumerate}
    \item \textbf{Mapping the Race Track}: The F1/10 racecar is manually driven around the race track to build a 2D occupancy grid map using the Hector SLAM algorithm~\cite{kohlbrecher2013hector}. 
    \item \textbf{ROSMap2Gazebo}: We extrude the map bounds by using a smoothing filter and export the resulting 3D mesh to Gazebo as a world model.
    \item \textbf{Label Assignment}: The lookahead label assignment algorithm is run on the virtual race track in ROS F1Tenth simulator for $\beta = [0.0, 0.25, 0.5, 0.75, 1.0]$, and the resulting lookahead label sets are benchmarked for performance.
    \item \textbf{Validation on Testbed}: The labels generated from our algorithm are exported to the F1/10 testbed and verified against simulation results.
\end{enumerate}
In doing so, we can go from a real track, to a real map, to a simulated track and back to the testbed (Fig.~\ref{fig:realworld}).

\subsection{Testbed Execution \& Results}


For accurate localization at high speeds, the F1/10 testbed was equipped with the CDDT particle filter using a GPU enabled ray-tracing algorithm~\cite{walsh17}.

In Fig.~\ref{fig:results_graphs}, the left half shows the reference trajectory imposed with the lookahead labels where read, yellow and green represent short, medium and long lookahead distances respectively and the chart showing the effect of the trade-off factor $\beta$ on the best lap time for the current setting. Note that the extreme emphasis on either velocity or deviation optimization leads to worse lap times as opposed to a balanced emphasis.

Observed lap times differences between simulation and real world implementation were within 0.5 seconds, and the total lap deviation during real world implementation was withing $5\%$ of the simulated deviation. 
This can be seen in the right half of Fig.~\ref{fig:results_graphs} which compares race metrics of the F1/10 autonomous racecar on the real race-track. The $convex\_combination$ label assignment has better performance in both lap time and average lap speed on the F1/10 testbed with $20\%$ improvement over the baseline implementation. 
The convex factor $\beta$ and its impact on the lap time is shown in Fig.~\ref{fig:results_graphs}.
As $\beta$ changes from 0.25 (minimum deviation) to 0.75 (maximum velocity), the label assignments produce a varying lap time with the best performance on all metrics at around $\beta$=0.5. At $\beta$=0.0, the racecar's performance was very similar to the baseline Ackermann-adjusted pure-pursuit, and several lookahead labels for $\beta$=1 led to undesirable behaviors including oscillations, drifting and general loss of path tracking on multiple turns.


\section{Conclusion \& Future Work}
In this paper we have demonstrated that adaptive lookahead pure-pursuit out performs Ackermann-steering adjusted pure-pursuit in terms of race related metrics such as lap time and average lap speed, and is a novel fit for autonomous racing, both in simulation and the F1/10 testbed. 
The analysis focuses on a single agent setting, where a single race car is tasked with following a reference trajectory with the minimum lap time. 
Our future work involves using the adaptive lookahead pure-pursuit for multiple autonomous racecars \& creating a formal framework for autonomous overtaking at high speeds and close-proximity situations.

\bibliographystyle{unsrt}
\bibliography{root.bbl}

\end{document}